# n-ary Fuzzy Logic and Neutrosophic Logic Operators

Florentin Smarandache, Chair of Math & Sc. Dept.
University of New Mexico, Gallup, NM 87301, USA, smarand@unm.edu

and
V. Christianto, SciPrint.org administrator, vxianto@yahoo.com

**Abstract**.

We extend Knuth's 16 Boolean binary logic operators to fuzzy logic and neutrosophic logic binary operators. Then we generalize them to n-ary fuzzy logic and neutrosophic logic operators using the smarandache codification of the Venn diagram and a defined vector neutrosophic law. In such way, new operators in neutrosophic logic/set/probability are built.

**Introduction**.

For the beginning let's consider the Venn Diagram of two variables $x$ and $y$, for each possible operator, as in Knuth's table, but we adjust this table to the Fuzzy Logic (FL).

Let's denote the fuzzy logic values of these variables as

$$FL(x) = (t_1, f_1)$$

where

$t_1$ = truth value of variable $x$,

$f_1$ = falsehood value of variable x,

with $0 \le t_1, f_1 \le 1$ and $t_1 + f_1 = 1$;

and similarly for $y$:

$$FL(y) = (t_2, f_2)$$

with the same $0 \le t_2, f_2 \le 1$ and $t_2 + f_2 = 1$.

We can define all 16 Fuzzy Logical Operators with respect to two $FL$ operators: $FL$ conjunction $(FLC)$ and $FL$ negation $(FLN)$.

Since in $FL$ the falsehood value is equal to 1- truth value, we can deal with only one component: the truth value.

The Venn Diagram for two sets $X$ and $Y$

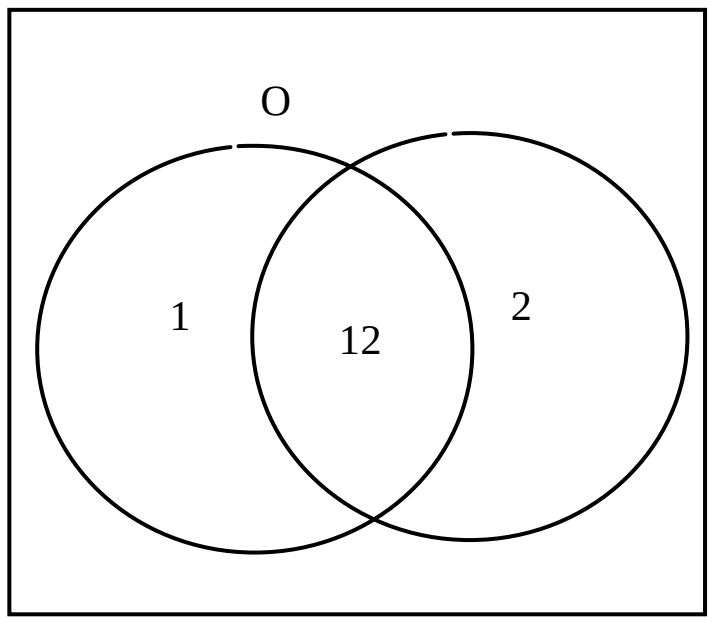

has $2^2 = 4$ disjoint parts:

0 = the part that does not belong to any set (the complement or negation)
1 = the part that belongs to 1st set only;
2 = the part that belongs to 2nd set only;
12 = the part that belongs to 1st and 2nd set only;
{called Smarandache's codification [1]}.

Shading none, one, two, three, or four parts in all possible combinations will make $2^4 = 2^{2^2} = 16$ possible binary operators.

We can start using a $T-norm$ and the negation operator.

Let's take the binary conjunction or intersection (which is a $T-norm$) denoted as $c_F(x,y)$:

$$c_F : \left([0,1]\times[0,1]\right)^2 \rightarrow [0,1]\times[0,1]$$

and unary negation operator denoted as $n_F(x)$, with:

$$n_F : [0,1]\times[0,1] \rightarrow [0,1]\times[0,1]$$

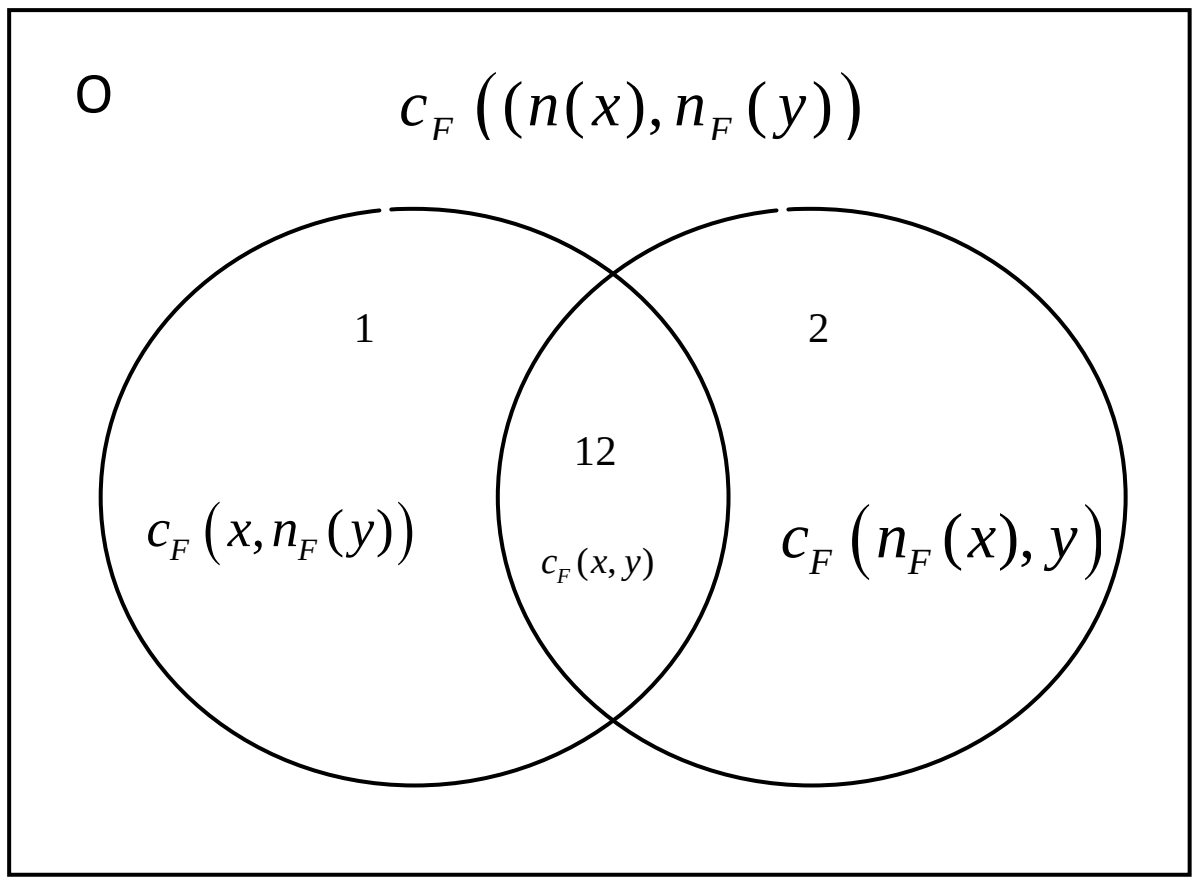

The fuzzy logic value of each part is:

$P12 = part12 =$ intersection of $x$ and $y$; so $FL(P12) = c_F(x,y)$.

$P1 = part1 =$ intersection of $x$ and negation of $y$; $FL(P1) = c_F(x, n_F(y))$.

$P2 = part2 =$ intersection of negation of $x$ and $y$; $FL(P2) = c_F(n_F(x), y)$.

$P0 = part0 =$ intersection of negation of $x$ and the negation of $y$; $FL(P0) = c_F(n_F(x), n_F(y))$, and for normalization we set the condition:

$$c_F(x,y) + c_F\left((n(x), y\right) + c_F\left(x, n_F(y)\right) + c_F\left(n_F(x), n_F(y)\right) = (1,0).$$

Then consider a binary $T-conorm$ (disjunction or union), denoted by $d_F(x,y)$:

$$d_F : \left([0,1]\times[0,1]\right)^2 \rightarrow [0,1]\times[0,1]$$

$$d_F\left(x,y\right) = \left(t_1+t_2,\ f_1+f_2-1\right)$$

if $x$ and $y$ are disjoint and $t_1+t_2 \le 1$.

This fuzzy disjunction operator of disjoint variables allows us to add the fuzzy truth-values of disjoint parts of a shaded area in the below table. When the truth-value increases, the false value decreases. More general, $d_F^k\left(x_1,x_2,...,x_k\right)$, as a k-ary disjunction (or union), for $k \ge 2$, is defined as:

$$d_F^k : \left([0,1]\times[0,1]\right)^k \rightarrow [0,1]\times[0,1]$$

$$d_F^k\left(x_1,x_2,...,x_k\right) = \left(t_1+t_2+...+t_k,\ f_1+f_2+...+f_k-k+1\right)$$

if all $x_i$ are disjoint two by two and $t_1+t_2+...+t_k \le 1$.

As a particular case let's take as a binary fuzzy conjunction:

$$c_F\left(x,y\right) = \left(t_1t_2,\ f_1+f_2-f_1f_2\right)$$

and as unary fuzzy negation:

$$n_F(x) = \left(1-t_1, 1-f_1\right) = \left(f_1, t_1\right),$$

where

$$FL(x) = (t_1,\ f_1),\ \text{with } t_1+f_1=1,\ \text{and } 0\le t_1,\ f_1\le 1;$$

$$FL(y) = (t_2,\ f_2),\ \text{with } t_2+f_2=1,\ \text{and } 0\le t_2,\ f_2\le 1.$$

whence:

$$FL(P12) = \left(t_1t_2,\ f_1+f_2-f_1f_2\right)$$

$$FL(P1) = \left(t_1f_2,\ f_1+t_2-f_1t_2\right)$$

$$FL(P2) = \left(f_1t_2, t_1+f_2-t_1f_2\right)$$

$$FL(P0) = \left(f_1f_2, t_1+t_2-t_1t_2\right)$$

The Venn Diagram for $n=2$ and considering only the truth values, becomes:

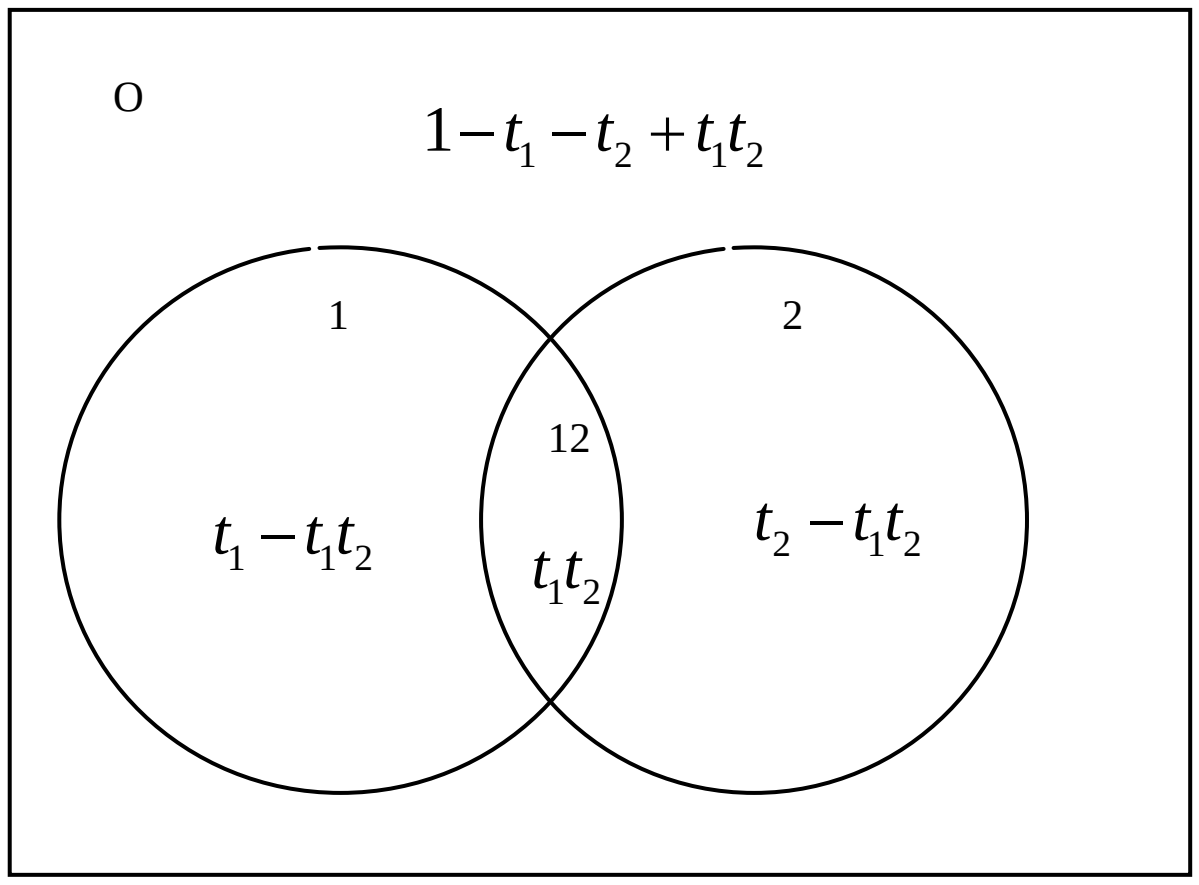

since

$$t_1 f_2 = t_1(1-t_2) = t_1 - t_1t_2$$
$$f_1t_2 = (1-t_1)t_2 = t_2 - t_1t_2$$
$$f_1 f_2 = (1-t_1)(1-t_2) = 1 - t_1 - t_2 + t_1t_2 .$$

We now use:

$$d_F^k\left(P12, P1, P2, P0\right) = \left(\left(t_1t_2\right) + \left(t_1 - t_1t_2\right) + \left(t_2 - t_1t_2\right) + \left(1 - t_1 - t_2 + t_1t_2\right),\right.$$
$$\left.\left(f_1 + f_2 - f_1 f_2\right) + \left(f_1 + t_2 - f_1t_2\right) + \left(t_1 + f_2 - t_1 f_2\right) + \left(t_1 + t_2 - t_1t_2\right) - 3\right) = \left(1, 0\right) .$$

So, the whole fuzzy space is normalized under $FL(\cdot)$ .

For the neurosophic logic, we consider

$NL(x) = (T_1, I_1, F_1)$, with $0 \le T_1, I_1, F_1 \le 1$;

$NL(y) = (T_2, I_2, F_2)$, with $0 \le T_2, I_2, F_2 \le 1$;

if the sum of components is 1 as in Atanassov's intuitionist fuzzy logic, i.e. $T_i + I_i + F_i = 1$, they are considered *normalized*; otherwise *non-normalized*, i.e. the sum of the components is <1 (*sub-normalized*) or >1 (*over-normalized*).

We define a binary neutrosophic conjunction (intersection) operator, which is a particular case of an N-norm (neutrosophic norm, a generalization of the fuzzy t-norm):

$$c_N : \left(\left[0,1\right]\times\left[0,1\right]\times\left[0,1\right]\right)^2 \to \left[0,1\right]\times\left[0,1\right]\times\left[0,1\right]$$

$$c_N(x, y) = \left(T_1T_2, I_1I_2 + I_1T_2 + T_1I_2, F_1F_2 + F_1I_2 + F_1T_2 + F_2T_1 + F_2I_1\right).$$

The neutrosophic conjunction (intersection) operator $x \wedge_N y$ component truth, indeterminacy, and falsehood values result from the multiplication

$$\left(T_1 + I_1 + F_1\right)\cdot\left(T_2 + I_2 + F_2\right)$$

since we consider in a prudent way $T \prec I \prec F$, where "$\prec$" means "weaker", i.e. the products $T_iI_j$ will go to $I$, $T_iF_j$ will go to $F$, and $I_iF_j$ will go to $F$ (or reciprocally we can say that $F$ prevails in front of $I$ and of $T$,

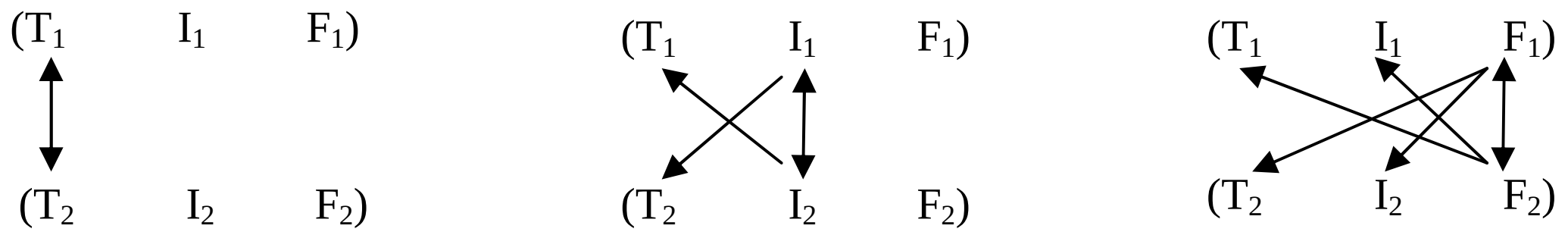

So, the truth value is $T_1T_2$, the indeterminacy value is $I_1I_2 + I_1T_2 + T_1I_2$ and the false value is $F_1F_2 + F_1I_2 + F_1T_2 + F_2T_1 + F_2I_1$. The norm of $x \wedge_N y$ is $\left(T_1 + I_1 + F_1\right)\cdot\left(T_2 + I_2 + F_2\right)$. Thus, if $x$ and $y$ are normalized, then $x \wedge_N y$ is also normalized. Of course, the reader can redefine the neutrosophic conjunction operator, depending on application, in a different way, for example in a more optimistic way, i.e. $I \prec T \prec F$ or $T$ prevails with respect to $I$, then we get:

$$c_N^{ITF}(x, y) = \left(T_1T_2 + T_1I_2 + T_2I_1, I_1I_2, F_1F_2 + F_1I_2 + F_1T_2 + F_2T_1 + F_2I_1\right).$$

Or, the reader can consider the order $T \prec F \prec I$, etc.

Let's also define the unary neutrosophic negation operator:

$$n_N : \left[0,1\right]\times\left[0,1\right]\times\left[0,1\right] \to \left[0,1\right]\times\left[0,1\right]\times\left[0,1\right]$$

$$n_N(T, I, F) = (F, I, T)$$

by interchanging the truth $T$ and falsehood $F$ vector components.
Then:

$$NL(P12) = (T_1T_2, I_1I_2 + I_1T_2 + I_2T_1, F_1F_2 + F_1I_2 + F_1T_2 + F_2T_1 + F_2I_1)$$

$$NL(P1) = (T_1F_2, I_1I_2 + I_1F_2 + I_2T_1, F_1T_2 + F_1I_2 + F_1F_2 + T_2T_1 + T_2I_1)$$

$$NL(P2) = (F_1T_2, I_1I_2 + I_1T_2 + I_2F_1, T_1F_2 + T_1I_2 + T_1T_2 + F_2F_1 + F_2I_1)$$

$$NL(P0) = (F_1F_2, I_1I_2 + I_1F_2 + I_2F_1, T_1T_2 + T_1I_2 + T_1F_2 + T_2F_1 + T_2I_1)$$

Similarly as in our above fuzzy logic work, we now define a binary $N-conorm$ (disjunction or union), i.e. neutrosophic conform.

$$d_N : ([0,1]\times[0,1]\times[0,1])^2 \to [0,1]\times[0,1]\times[0,1]$$

$$d_N(x, y) = \left(T_1 + T_2, (I_1 + I_2)\cdot\frac{\tau - T_1 - T_2}{I_1 + I_2 + F_1 + F_2}, (F_1 + F)\cdot\frac{\tau - T_1 - T_2}{I_1 + I_2 + F_1 + F_2}\right)$$

if $x$ and $y$ are disjoint, and $T_1 + T_2 \le 1$ where $\tau$ is the neutrosophic norm of $x \vee_N y$, i.e.

$$\tau = (T_1 + I_1 + F_1)\cdot(T_2 + I_2 + F_2).$$

We consider as neutrosophic norm of $x$, where $NL(x) = T_1 + I_1 + F_1$, the sum of its components: $T_1 + I_1 + F_1$, which in many cases is 1, but can also be positive <1 or >1.

When the truth value increases $(T_1 + T_2)$ is the above definition, the indeterminacy and falsehood values decrease proportionally with respect to their sums $I_1 + I_2$ and respectively $F_1 + F_2$.

This neutrosophic disjunction operator of disjoint variables allows us to add neutrosophic truth values of disjoint parts of a shaded area in a Venn Diagram.

Now, we complete Donald E. Knuth's Table of the Sixteen Logical Operators on two variables with Fuzzy Logical operators on two variables with Fuzzy Logic truth values, and Neutrosophic Logic truth/indeterminacy/false values (for the case $T \prec I \prec F$).

Table 1

| **Fuzzy Logic Truth Values** | **Venn Diagram** | **Notations** | **Operator symbol** | **Name(s)** |
|---|---|---|---|---|
| $0$ | 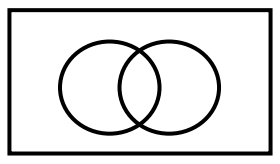 | $0$ | $\bot$ | Contradiction, falsehood; constant 0 |
| $t_1 t_2$ | 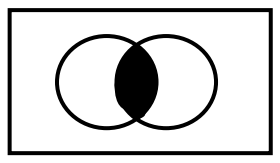 | $xy,\ x \wedge y,\ x \,\&\, y$ | $\wedge$ | Conjunction; and |
| $t_1 - t_1 t_2$ | 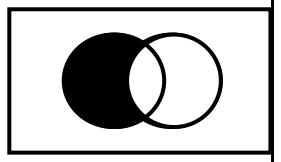 | $x \wedge \bar{y}, x \not\supset y, [x > y], x - y$ | $\bar{\supset}$ | Nonimplication; difference, but not |
| $t_1$ | 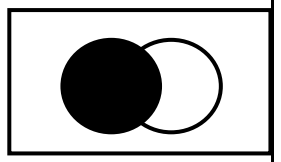 | $x$ | $L$ | Left projection |
| $t_2 - t_1 t_2$ | 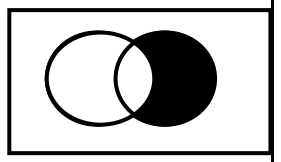 | $\bar{x} \wedge y, x \not\subset y, [x < y], y - x$ | $\bar{\subset}$ | Converse nonimplication; not...but |
| $t_2$ | 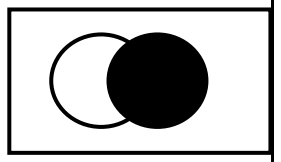 | $y$ | $R$ | Right projection |
| $t_1 + t_2 - 2t_1 t_2$ | 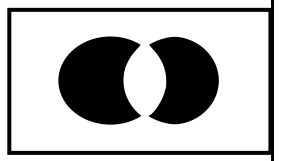 | $x \oplus y, x \not\equiv y, x^{\wedge} y$ | $\oplus$ | Exclusive disjunction; nonequivalence; "xor" |
| $t_1 + t_2 - t_1 t_2$ | 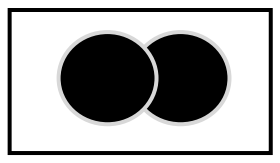 | $x \vee y, x \mid y$ | $\vee$ | (Inclusive) disjunction; or; and/or |
| $1 - t_1 - t_2 + t_1 t_2$ | 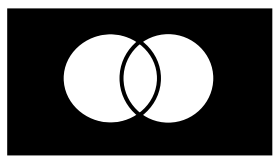 | $\bar{x} \wedge \bar{y}, \overline{x \vee y}, x \bar{\vee} y, x \uparrow y$ | $\bar{\vee}$ | Nondisjunction, joint denial, neither...nor |
| $1 - t_1 - t_2 + 2t_1 t_2$ | 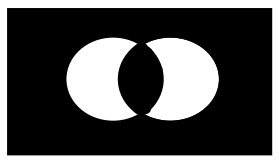 | $x \equiv y, x \leftrightarrow y, x \Leftrightarrow y$ | $\equiv$ | Equivalence; if and only if |
| $1 - t_2$ | 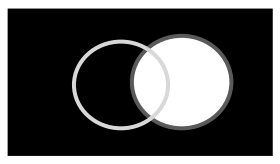 | $\bar{y}, \neg y, !y, \sim y$ | $\bar{R}$ | Right complementation |
| $1 - t_2 + t_1 t_2$ | 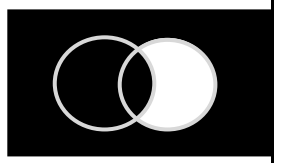 | $x \vee \bar{y}, x \subset y, x \Leftarrow y,$ $[x \geq y], x^y$ | $\subset$ | Converse implication if |
| $1 - t_1$ | 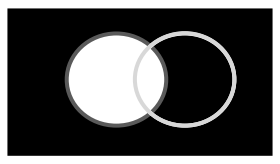 | $\bar{x}, \neg x, !x, \sim x$ | $\bar{L}$ | Left complementation |
| $1 - t_1 + t_1 t_2$ | 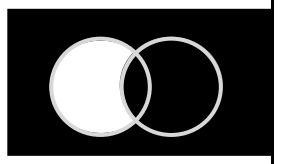 | $\bar{x} \vee y, x \supset y, x \Rightarrow y,$ $[x \leq y], y^x$ | $\supset$ | Implication; only if; if..then |
| $1-t_1t_2$ | 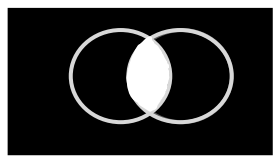 | $\overline{x}\vee\overline{y}, \overline{x\wedge y}, x\overline{\wedge}\, y, x\mid y$ | $\overline{\wedge}$ | Nonconjunction, not both…and; “nand” |
| $1$ | 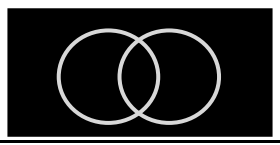 | $1$ | $T$ | Affirmation; validity; tautology; constant 1 |

Table 2

| Venn Diagram | Neutrosophic Logic Values |
|---|---|
| 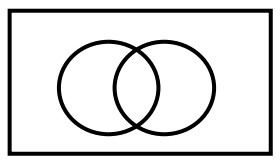 | $(0,0,1)$ |
| 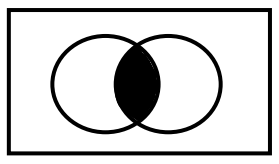 | $(T_1T_2, I_1I_2 + IT, F_1F_2 + FI + FT)$, where $IT = I_1T_2 + I_2T_1$ similarly $FI, FT$ ; |
| 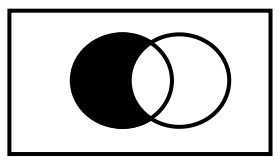 | $\left(T_1F_2, \underbrace{I_1I_2 + IT_{\overline{y}}}_{I_{P1}}, \underbrace{F_{\overline{y}}F_{\overline{y}} + F_{\overline{y}}I + F_{\overline{y}}T}_{F_{P2}}\right)$ |
| 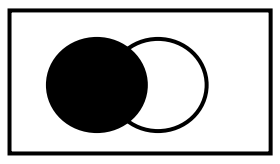 | $(T_1, I_1, F_1)$ |
| 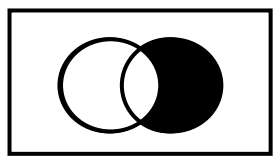 | $\left(F_1T_2, \underbrace{I_1I_2 + IT_{\overline{x}}}_{I_{P2}}, \underbrace{F_{\overline{x}}F_{\overline{x}} + F_{\overline{x}}I + F_{\overline{x}}T}_{F_{P2}}\right)$ |
| 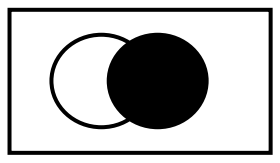 | $(T_2, I_2, F_2)$ |
| 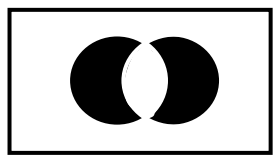 | $\left(TF, (I_{P1} + I_{P2})\cdot\frac{\tau - TF}{I_{P1} + I_{P2} + F_{P1} + F_{P2}}, (F_{P1} + F_{P2})\cdot\frac{\tau - TF}{I_{P1} + I_{P2} + F_{P1} + F_{P2}}\right)$ Where $\tau = (T_1 + I_1 + F_1)\cdot(T_2 + I_2 + F_2)$ which is the neutrosophic norm |
| 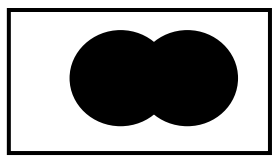 | $(T_1T_2 + TI + TF, I_1I_2 + IF, F_1F_2)$ |
| 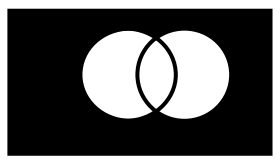 | $(F_1F_2, I_1I_2 + IF, T_1T_2 + TI + TF)$ |
| 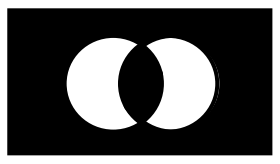 | $\left((F_{P1} + F_{P2})\cdot\frac{\tau - TF}{I_{P1} + I_{P2} + F_{P1} + F_{P2}}, (I_{P1} + I_{P2})\cdot\frac{\tau - TF}{I_{P1} + I_{P2} + F_{P1} + F_{P2}}, TF\right)$ |
| 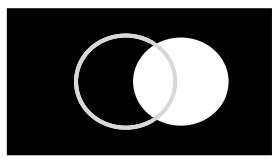 | $(F_2, I_2, T_2)$ |
| 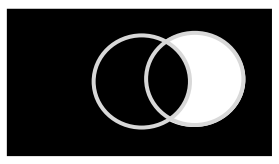 | $(F_{\overline{x}}F_{\overline{x}} + F_{\overline{x}}I + F_{\overline{x}}T, I_1I_2 + IT_{\overline{x}}, F_1T_2)$ |
| 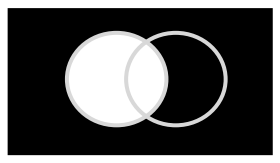 | $(F_1, I_1, T_1)$ |
| 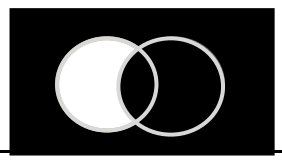 | $(F_{\overline{y}}F_{\overline{y}} + F_{\overline{y}}I + F_{\overline{y}}T, I_1I_2 + IT_{\overline{y}}, T_1F_2)$ |
| 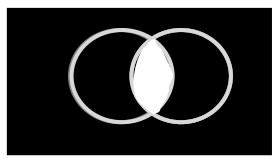 | $\left(F_1F_2 + FI + FT, I_1I_2 + IT, T_1T_2\right)$ |
| 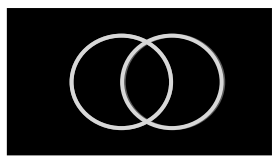 | $(1,0,0)$ |

These 16 neutrosophic binary operators are approximated, since the binary N-conorm gives an approximation because of 'indeterminacy' component.

### Tri-nary Fuzzy Logic and Neutrosophic Logic Operators

In a more general way, for $k \geq 2$:

$$d_N^k : \left([0,1]\times[0,1]\times[0,1]\right)^k \to [0,1]\times[0,1]\times[0,1],$$

$$d_N^k\left(x_1, x_2, ..., x_k\right) = \left( \sum_{i=1}^{k} T_i, \left(\sum_{i=1}^{k} I_i\right) \cdot \frac{\tau_k - \sum_{i=1}^{k} T_i}{\sum_{i=1}^{k}\left(I_i + F_i\right)}, \left(\sum_{i=1}^{k} F_i\right) \cdot \frac{\tau_k - \sum_{i=1}^{k} T_i}{\sum_{i=1}^{k}\left(I_i + F_i\right)} \right)$$

if all $x_i$ are disjoint two by two, and $\sum_{i=1}^{k} T_i \leq 1$.

We can extend Knuth's Table from binary operators to tri-nary operators (and we get $2^{2^3} = 256$ tri-nary operators) and in general to n-ary operators (and we get $2^{2^n}$ n-ary operators).

Let's present the tri-nary Venn Diagram, with 3 variables $x, y, z$

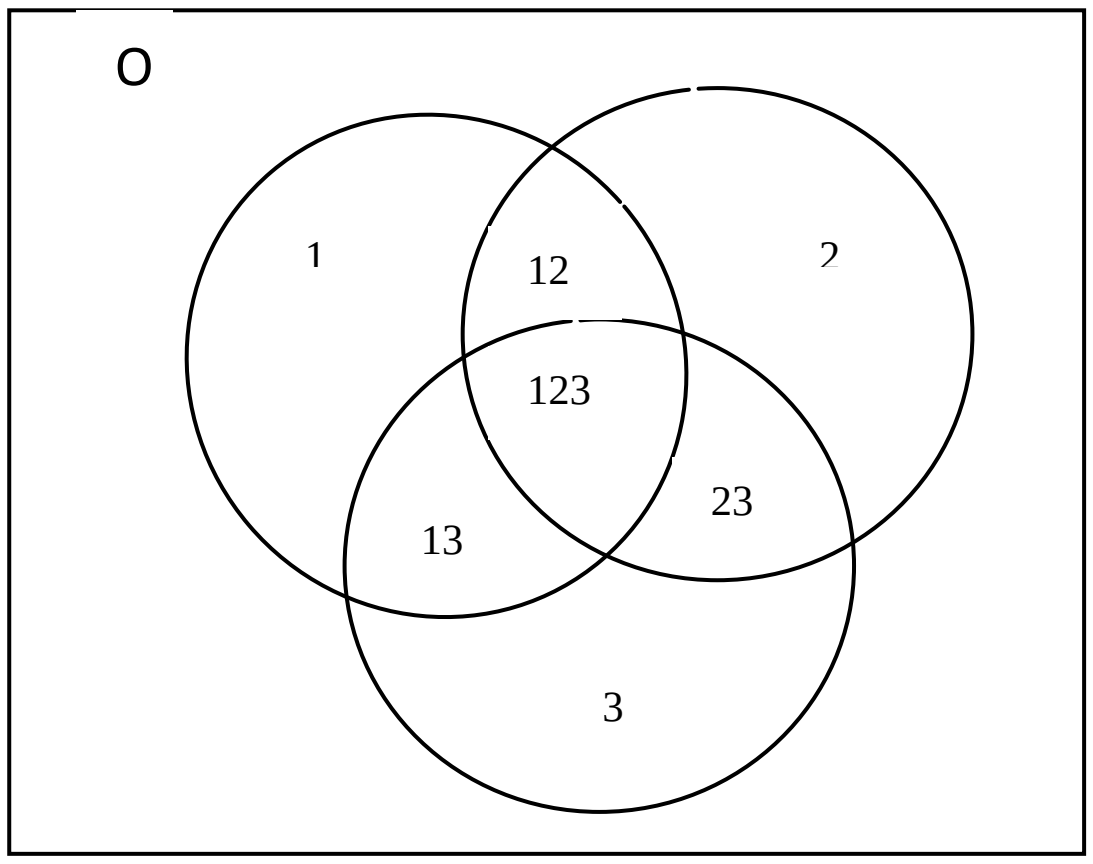

using the name Smarandache codification.

This has $2^3 = 8$ disjoint parts, and if we shade none, one, two, …, or eight of them and consider all possible combinations we get $2^8 = 256$ tri-nary operators in the above tri-nary Venn Diagram.

For n=3 we have:

$$P123 = c_F(x, y, z)$$
$$P12 = c_F\left(x, y, n_F(z)\right)$$
$$P13 = c_F\left(x, n_F(y), z\right)$$
$$P23 = c_F\left(n_F(x), y, z\right)$$
$$P1 = c_F\left(x, n_F(y), n_F(z)\right)$$
$$P2 = c_F\left(n_F(x), y, n_F(z)\right)$$
$$P3 = c_F\left(n_F(x), n_F(y), z\right)$$
$$P0 = c_F\left(n_F(x), n_F(y), n_F(z)\right)$$

Let

$FL(x) = (t_1, f_1)$, with $t_1 + f_1 = 1,\ \ 0 \le t_1, f_1 \le 1$,

$FL(y) = (t_2, f_2)$, with $t_2 + f_2 = 1,\ \ 0 \le t_2, f_2 \le 1$,

$FL(z) = (t_3, f_3)$, with $t_3 + f_3 = 1,\ \ 0 \le t_3, f_3 \le 1$.

We consider the particular case defined by tri-nary conjunction fuzzy operator:

$$c_F : \left([0,1]\times[0,1]\right)^3 \to [0,1]\times[0,1]$$
$$c_F(x, y, z) = \left(t_1t_2t_3,\ f_1 + f_2 + f_3 - f_1 f_2 - f_2 f_3 - f_3 f_1 + f_1 f_2 f_3\right)$$

because

$$\left((t_1, f_1)\wedge_F (t_2, f_2)\right)\wedge_F (t_3, f_3) = \left(t_1t_2,\ f_1 + f_2 - f_1 f_2\right)\wedge_F (t_3, f_3) =$$
$$= \left(t_1t_2t_3,\ f_1 + f_2 + f_3 - f_1 f_2 - f_2 f_3 - f_3 f_1 + f_1 f_2 f_3\right)$$

and the unary negation operator:

$$n_F : \left([0,1]\times[0,1]\right) \to [0,1]\times[0,1]$$
$$n_F(x) = (1 - t_1, 1 - f_1) = (f_1, t_1).$$

We define the function:

$$L_1 : [0,1]\times[0,1]\times[0,1] \to [0,1]$$
$$L_1(\alpha, \beta, \gamma) = \alpha\cdot\beta\cdot\gamma$$

and the function

$$L_2 : [0,1]\times[0,1]\times[0,1] \to [0,1]$$
$$L_2(\alpha, \beta, \gamma) = \alpha + \beta + \gamma - \alpha\beta - \beta\gamma - \gamma\alpha + \alpha\beta\gamma$$

then:

$$FL(P123) = \left(L_1(t_1, t_2, t_3), L_2(f_1, f_2, f_3)\right)$$
$$FL(P12) = \left(L_1(t_1, t_2, f_3), L_2(f_1, f_2, t_3)\right)$$
$$FL(P13) = \left(L_1(t_1, f_2, t_3), L_2(f_1, t_2, f_3)\right)$$
$$FL(P23) = \left(L_1(f_1, t_2, t_3), L_2(t_1, f_2, f_3)\right)$$
$$FL(P1) = \left(L_1(t_1, f_2, f_3), L_2(f_1, t_2, t_3)\right)$$
$$FL(P2) = \left(L_1(f_1, t_2, f_3), L_2(t_1, f_2, t_3)\right)$$
$$FL(P3) = \left(L_1(f_1, f_2, t_3), L_2(t_1, t_2, f_3)\right)$$
$$FL(P0) = \left(L_1(f_1, f_2, f_3), L_2(t_1, t_2, t_3)\right)$$

We thus get the fuzzy truth-values as follows:

$$FL_t(P123) = t_1 t_2 t_3$$
$$FL_t(P12) = t_1 t_2 (1 - t_3) = t_1 t_2 - t_1 t_2 t_3$$
$$FL_t(P13) = t_1 (1 - t_2) t_3 = t_1 t_3 - t_1 t_2 t_3$$
$$FL_t(P23) = (1 - t_1) t_2 t_3 = t_2 t_3 - t_1 t_2 t_3$$
$$FL_t(P1) = t_1 (1 - t_2)(1 - t_3) = t_1 - t_1 t_2 - t_1 t_3 + t_1 t_2 t_3$$
$$FL_t(P2) = (1 - t_1) t_2 (1 - t_3) = t_2 - t_1 t_2 - t_2 t_3 + t_1 t_2 t_3$$
$$FL_t(P3) = (1 - t_1)(1 - t_2) t_3 = t_3 - t_1 t_3 - t_2 t_3 + t_1 t_2 t_3$$
$$FL_t(P0) = (1 - t_1)(1 - t_2)(1 - t_3) = 1 - t_1 - t_2 - t_3 + t_1 t_2 + t_1 t_3 + t_2 t_3 - t_1 t_2 t_3 .$$

We, then, consider the same disjunction or union operator $d_F(x, y) = t_1 + t_2, f_1 + f_2 - 1$, if $x$ and $y$ are disjoint, and $t_1 + t_2 \leq 1$ allowing us to add the fuzzy truth values of each part of a shaded area.

## Neutrophic Composition Law

Let's consider $k \geq 2$ neutrophic variables, $x_i\left(T_i, I_i, F_i\right)$, for all $i \in \left\{1, 2, ..., k\right\}$. Let denote

$$T = \left(T_1, ..., T_k\right)$$
$$I = \left(I_1, ..., I_k\right)$$
$$F = \left(F_1, ..., F_k\right) .$$

We now define a neutrosophic composition law $o_N$ in the following way:

$$o_N : \left\{T, I, F\right\} \to \left[0, 1\right]$$

If $z \in \left\{T, I, F\right\}$ then $z_{o_N} z = \prod_{i=1}^{k} z_i$ .

If $z, w \in \left\{T, I, F\right\}$ then

$$z_{o_N} w = w_{o_N} z \quad = \sum_{\substack{r=1 \\ \{i_1,...,i_r,j_{r+1},...,j_k\}\equiv\{1,2,...,k\} \\ (i_1,...,i_r)\in C^r(1,2,...,k) \\ (j_{r+1},...,j_k)\in C^{k-r}(1,2,...,k)}}^{k-1} z_{i_1}...z_{i_r} w_{j_{r+1}}...w_{j_k}$$

where $C^r(1,2,...,k)$ means the set of combinations of the elements $\{1,2,...,k\}$ taken by $r$. [Similarly for $C^{k-r}(1,2,...,k)$].

In other words, $z_{o_N} w$ is the sum of all possible products of the components of vectors $z$ and $w$, such that each product has at least a $z_i$ factor and at least $w_j$ factor, and each product has exactly $k$ factors where each factor is a different vector component of $z$ or of $w$. Similarly if we multiply three vectors:

$$T_{o_N} I_{o_N} F = \sum_{\substack{u,v,k-u-v=1 \\ \{i_1,...,i_u,j_{u+1},...,j_{u+v},l_{u+v+1},...,l_k\}\equiv\{1,2,...,k\} \\ (i_1,...,i_u)\in C^u(1,2,...,k),(j_{u+1},...,j_{u+v})\in \\ \in C^v(1,2,...,k),(l_{u+v+1},...,l_k)\in C^{k-u-v}(1,2,...,k)}}^{k-2} T_{i_1...i_u} I_{j_{u+1}...j_{u+v}} F_{l_{u+v+1}}...F_{l_k}$$

Let's see an example for $k=3$.

$$x_1(T_1, I_1, F_1)$$
$$x_2(T_2, I_2, F_2)$$
$$x_3(T_3, I_3, F_3)$$

$$T_{o_N} T = T_1T_2T_3, \quad I_{o_N} I = I_1I_2I_3, \quad F_{o_N} F = F_1F_2F_3$$
$$T_{o_N} I = T_1I_2I_3 + I_1T_2I_3 + I_1I_2T_3 + T_1T_2I_3 + T_1I_2T_3 + I_1T_2T_3$$
$$T_{o_N} F = T_1F_2F_3 + F_1T_2F_3 + F_1F_2T_3 + T_1T_2F_3 + T_1F_2T_3 + F_1T_2T_3$$
$$I_{o_N} F = I_1F_2F_3 + F_1I_2F_3 + F_1F_2I_3 + I_1I_2F_3 + I_1F_2I_3 + F_1I_2I_3$$
$$T_{o_N} I_{o_N} F = T_1I_2F_3 + T_1F_2I_3 + I_1T_2F_3 + I_1F_2T_3 + F_1I_2T_3 + F_1T_2I_3$$

For the case when indeterminacy $I$ is not decomposed in subcomponents {as for example $I = P \cup U$ where $P$ =paradox (true and false simultaneously) and $U$ =uncertainty (true or false, not sure which one)}, the previous formulas can be easily written using only three components as:

$$T_{o_N} I_{o_N} F = \sum_{i,j,r\in\mathcal{P}(1,2,3)} T_i I_j F_r$$

where $\mathcal{P}(1,2,3)$ means the set of permutations of $(1,2,3)$ i.e.

$$\{(1,2,3),(1,3,2),(2,1,3),(2,3,1,),(3,1,2),(3,2,1)\}$$

$$z_{o_N} w = \sum_{\substack{\mathrm{i}=1 \\ (i,j,r)\equiv(1,2,3) \\ (j,r)\in\mathcal{P}^2(1,2,3)}}^{3} z_i w_j w_{j_r} + w_i z_j z_r$$

This neurotrophic law is associative and commutative.

## Neutrophic Logic Operators

Let's consider the neutrophic logic cricy values of variables $x, y, z$ (so, for $n=3$) $NL(x)=(T_1, I_1, F_1)$ with $0 \le T_1, I_1, F_1 \le 1$

$$NL(y)=(T_2, I_2, F_2) \text{ with } 0 \le T_2, I_2, F_2 \le 1$$

$$NL(z)=(T_3, I_3, F_3) \text{ with } 0 \le T_3, I_3, F_3 \le 1$$

In neutrosophic logic it is not necessary to have the sum of components equals to 1, as in intuitionist fuzzy logic, i.e. $T_k + I_k + F_k$ is not necessary 1, for $1 \le k \le 3$

As a particular case, we define the tri-nary conjunction neutrosophic operator:

$$c_N : ([0,1]\times[0,1]\times[0,1])^3 \to [0,1]\times[0,1]\times[0,1]$$

$$c_N(x,y)=\left(T_{o_N} T, I_{o_N} I + I_{o_N} T, F_{o_N} F + F_{o_N} I + F_{o_N} T\right)$$

If $x$ or $y$ are normalized, then $c_N(x,y)$ is also normalized.

If $x$ or $y$ are non-normalized then $|c_N(x,y)| = |x|\cdot|y|$ where $|\cdot|$ means norm.

$c_N$ is an N-norm (neutrosophic norm, i.e. generalization of the fuzzy t-norm).

Again, as a particular case, we define the unary negation neutrosophic operator:

$$n_N : [0,1]\times[0,1]\times[0,1] \to [0,1]\times[0,1]\times[0,1]$$

$$n_N(x) = n_N(T_1, I_1, F_1) = (F_1, I_1, T_1).$$

We take the same Venn Diagram for $n=3$.

So,

$$NL(x)=(T_1, I_1, F_1)$$

$$NL(y)=(T_2, I_2, F_2)$$

$$NL(z)=(T_3, I_3, F_3).$$

Vectors

$$\text{T}=\begin{pmatrix} T_1 \\ T_2 \\ T_3 \end{pmatrix}, \quad \text{I}=\begin{pmatrix} I_1 \\ I_2 \\ I_3 \end{pmatrix} \text{ and F}=\begin{pmatrix} F_1 \\ F_2 \\ F_3 \end{pmatrix}.$$

We note $\text{T}_{\bar{x}}=\begin{pmatrix} F_1 \\ T_2 \\ T_3 \end{pmatrix}$, $\text{T}_{\bar{y}}=\begin{pmatrix} T_1 \\ F_2 \\ T_3 \end{pmatrix}$, $\text{T}_{\bar{z}}=\begin{pmatrix} T_1 \\ T_2 \\ F_3 \end{pmatrix}$, $\text{T}_{\overline{xy}}=\begin{pmatrix} F_1 \\ F_2 \\ T_3 \end{pmatrix}$, etc.

and similarly

$$\text{F}_{\bar{x}}=\begin{pmatrix} T_1 \\ F_2 \\ F_3 \end{pmatrix}, \quad F_{\bar{y}}=\begin{pmatrix} F_1 \\ T_2 \\ F_3 \end{pmatrix}, \quad \text{F}_{\overline{xz}}=\begin{pmatrix} T_1 \\ F_2 \\ T_3 \end{pmatrix}, \text{ etc.}$$

For shorter and easier notations let's denote $z_{o_N} w = zw$ and respectively $z_{o_N} w_{o_N} v = zwv$ for the vector neutrosophic law defined previously.

Then

$$NL(P123) = c_N(x,y) = (TT, II + IT, FF + FI + FT + FIT) =$$

$$
\begin{aligned}
&=\left(T_1 T_2 T_3, I_1 I_2 I_3+I_1 I_2 T_3+I_1 T_2 I_3+T_1 I_2 I_3+I_1 T_2 T_3+T_1 I_2 T_3+T_1 T_2 I_3,\right.\\
&F_1 F_2 F_3+F_1 F_2 I_3+F_1 I_2 F_3+I_1 F_2 F_3+F_1 I_2 I_3+I_1 F_2 I_3+I_1 I_2 F_3+\\
&+F_1 F_2 T_3+F_1 T_2 F_3+T_1 F_2 F_3+F_1 T_2 T_3+T_1 F_2 T_3+T_1 T_2 F_3+\\
&\left.+T_1 I_2 F_3+T_1 F_2 I_3+I_1 F_2 T_3+I_1 T_2 F_3+F_1 I_2 T_3+F_1 T_2\, I_3\right)
\end{aligned}
$$

$$NL(P12)=c_N\left(x,y,n_N(z)\right)=\left(T_{\bar z}T_{\bar z}, II+IT_{\bar z}, F_{\bar z}F_{\bar z}+F_{\bar z}I+F_{\bar z}T_{\bar z}+F_{\bar z}IT_{\bar z}\right)$$

$$NL(P13)=c_N\left(x,n_N(y),z\right)=\left(T_{\bar y}T_{\bar y}, II+IT_{\bar y}, F_{\bar y}F_{\bar y}+F_{\bar y}I+F_{\bar y}T_{\bar y}+F_{\bar y}IT_{\bar y}\right)$$

$$NL(P23)=c_N\left(n_N(x),y,z\right)=\left(T_{\bar x}T_{\bar x}, II+IT_{\bar x}, F_{\bar x}F_{\bar x}+F_{\bar x}I+F_{\bar x}T_{\bar x}+F_{\bar x}IT_{\bar y}\right)$$

$$NL(P1)=c_N\left(x,n_N(y),n_N(z)\right)=\left(T_{\overline{yz}}T_{\overline{yz}}, II+IT_{\overline{yz}}, F_{\overline{yz}}F_{\overline{yz}}+F_{\overline{yz}}I+F_{\overline{yz}}T_{\overline{yz}}+F_{\overline{yz}}IT_{\overline{yz}}\right)$$

$$NL(P2)=c_N\left(n_N(x),y,n_N(z)\right)=\left(T_{\overline{xz}}T_{\overline{xz}}, II+IT_{\overline{xz}}, F_{\overline{xz}}F_{\overline{xz}}+F_{\overline{xz}}I+F_{\overline{xz}}T_{\overline{xz}}+F_{\overline{xz}}IT_{\overline{xz}}\right)$$

$$NL(P0)=c_N\left(n_N(x),n_N(y),n_N(z)\right)=\left(T_{\overline{xyz}}T_{\overline{xyz}}, II+IT_{\overline{xyz}}, F_{\overline{xyz}}F_{\overline{xyz}}+F_{\overline{xyz}}I+F_{\overline{xyz}}T_{\overline{xyz}}+F_{\overline{xyz}}IT_{\overline{xyz}}\right)=$$
$$=\left(FF, II+IF, TT+TI+TF+TIF\right).$$

## n-ary Fuzzy Logic and Neutrosophic Logic Operators

We can generalize for any integer $n \geq 2$.

The Venn Diagram has $2^{2^n}$ disjoint parts. Each part has the form $Pi_1 ... i_k j_{k+1} ... j_n$ , where $0 \leq k \leq n$ , and of course $0 \leq n-k \leq n$; $\{i_1,...,i_k\}$ is a combination of $k$ elements of the set $\{1,2,...,n\}$, while $\{j_{k+1},...,j_n\}$ the $n-k$ elements left, i.e. $\{j_{k+1},...,j_k\}=\{1,2,...,n\}\setminus\{i_1,...,i_k\}$. $\{i_1,...,i_k\}$ are replaced by the corresponding numbers from $\{1,2,...,n\}$, while $\{j_{k+1},...,j_n\}$ are replaced by blanks.

For example, when $n=3$,

$$Pi_1 i_2 j_3 = P13 \text{ if } \{i_1,i_2\}=\{1,3\},$$
$$Pi_1 j_2 j_3 = P1 \text{ if } \{i_1\}=\{1\}.$$

Hence, for fuzzy logic we have:

$$Pi_1...i_k j_{k+1}...j_n = c_F\left(x_{i_1},...,x_{i_k}, n_F\left(x_{j_{k+1}}\right),...,n_F\left(x_{j_n}\right)\right)$$

whence

$$FL\left(Pi_1...i_k j_{k+1}...j_n\right)=\left(\left(\prod_{r=1}^{k} t_{i_r}\right)\left(\prod_{s=k+1}^{n}\left(1-t_{j_s}\right)\right), \varphi\left(f_1 f_2,..., f_n\right)\right)$$

where $\varphi:[0,1]^n \to [0,1]$,

$$\varphi\left(\alpha_1,\alpha_2,...,\alpha_n\right)=S_1-S_2+S_3+...+(-1)^{n+1}S_n=\sum_{l=1}^{n}(-1)^{l+1}S_l$$

where

$$S_1 = \sum_{i=1}^{n} \alpha_i$$

$$S_2 = \sum_{1 \le i < j \le n} \alpha_i \alpha_j$$

.......................................

$$S_l = \sum_{1 \le i_1 < i_2 < \ldots < i_l \le n} \alpha_{i_1} \alpha_{i_2} \ldots \alpha_{i_l}$$

.......................................

$$S_n = \alpha_1 \cdot \alpha_2 \cdot \ldots \cdot \alpha_n$$

And for neutrosophic logic we have:

$$Pi_1 \ldots i_k j_{k+1} \ldots j_n = c_N \left( x_{i_1}, \ldots, x_{i_k}, n_N \left( x_{j_{k+1}} \right), \ldots, n_N \left( x_{j_n} \right) \right)$$

whence:

$$NL\left( Pi_1 \ldots i_k j_{k+1} \ldots j_n \right) = \left( T_{12\ldots n}, I_{12\ldots n}, F_{12\ldots n} \right),$$

where

$$T_{12\ldots n} = T_{\bar{x}_{j_{k+1}} \ldots \bar{x}_{j_n}} T_{\bar{x}_{j_{k+1}} \ldots \bar{x}_{j_n}} = \left( \prod_{r=1}^{k} T_{i_r} \right) \cdot \prod_{s=k+1}^{n} F_{j_s} .$$

$$I_{12\ldots n} = II + IT_{\bar{x}_{j_{k+1}} \ldots \bar{x}_{j_n}} ,$$

$$F_{12\ldots n} = F_{\bar{x}_{j_{k+1}} \ldots \bar{x}_{j_n}} F_{\bar{x}_{j_{k+1}} \ldots \bar{x}_{j_n}} + F_{\bar{x}_{j_{k+1}} \ldots \bar{x}_{j_n}} I + F_{\bar{x}_{j_{k+1}} \ldots \bar{x}_{j_n}} T_{\bar{x}_{j_{k+1}} \ldots \bar{x}_{j_n}} + F_{\bar{x}_{j_{k+1}} \ldots \bar{x}_{j_n}} IT_{\bar{x}_{j_{k+1}} \ldots \bar{x}_{j_n}}$$

**Conclusion:**

A generalization of Knuth's Boolean binary operations is presented in this paper, i.e. we present n-ary Fuzzy Logic Operators and Neutrosophic Logic Operators based on Smarandache's codification of the Venn Diagram and on a defined vector neutrosophic law which helps in calculating fuzzy and neutrosophic operators.

Better neutrosophic operators than in [2] are proposed herein.